\PassOptionsToPackage{prologue,dvipsnames,dvipsnames,table}{xcolor}

\documentclass[10pt,twocolumn,letterpaper]{article}

\usepackage[pagenumbers]{cvpr} %

\definecolor{cvprblue}{rgb}{0.21,0.49,0.74}
\usepackage[pagebackref,breaklinks,colorlinks,allcolors=cvprblue]{hyperref}
\usepackage{multirow}
\usepackage{gensymb}
\usepackage{amsmath}
\usepackage{xspace}
\usepackage{graphicx}
\usepackage[dvipsnames]{xcolor}
\usepackage[table]{xcolor}

\def\paperID{13674} %
\def\confName{CVPR}
\def\confYear{2025}

\newcommand{\RTA}{\mathrm{RTA}}
\newcommand{\RRA}{\mathrm{RRA}}

\newcommand{\oneptsmaller}[1]{%
  \begingroup
  \fontsize{\dimexpr\f@size pt-1pt}{\f@baselineskip}\selectfont
  #1%
  \endgroup
}

\definecolor{mygreen}{RGB}{0, 200, 0}
\definecolor{myorange}{RGB}{254, 178, 76}
\newcommand{\MYhref}[3][0000AA]{\href{#2}{\color[HTML]{#1}{\textbf{#3}}}}%

\title{AerialMegaDepth: Learning Aerial-Ground Reconstruction and View Synthesis}

\author{
Khiem Vuong \hspace{3em} Anurag Ghosh \\ Deva Ramanan$^{\ast}$ \quad Srinivasa Narasimhan$^{\ast}$ \quad Shubham Tulsiani$^{\ast}$ \vspace{4mm} \\
Carnegie Mellon University \\
{\normalsize \MYhref{https://aerial-megadepth.github.io}{\texttt{https://aerial-megadepth.github.io}}}\\
}

\begin{document}

\twocolumn[{
\renewcommand\twocolumn[1][]{#1}%
\maketitle
\vspace{-1cm}
\begin{center}
    \centering
    \captionsetup{type=figure}
    \includegraphics[width=\textwidth]{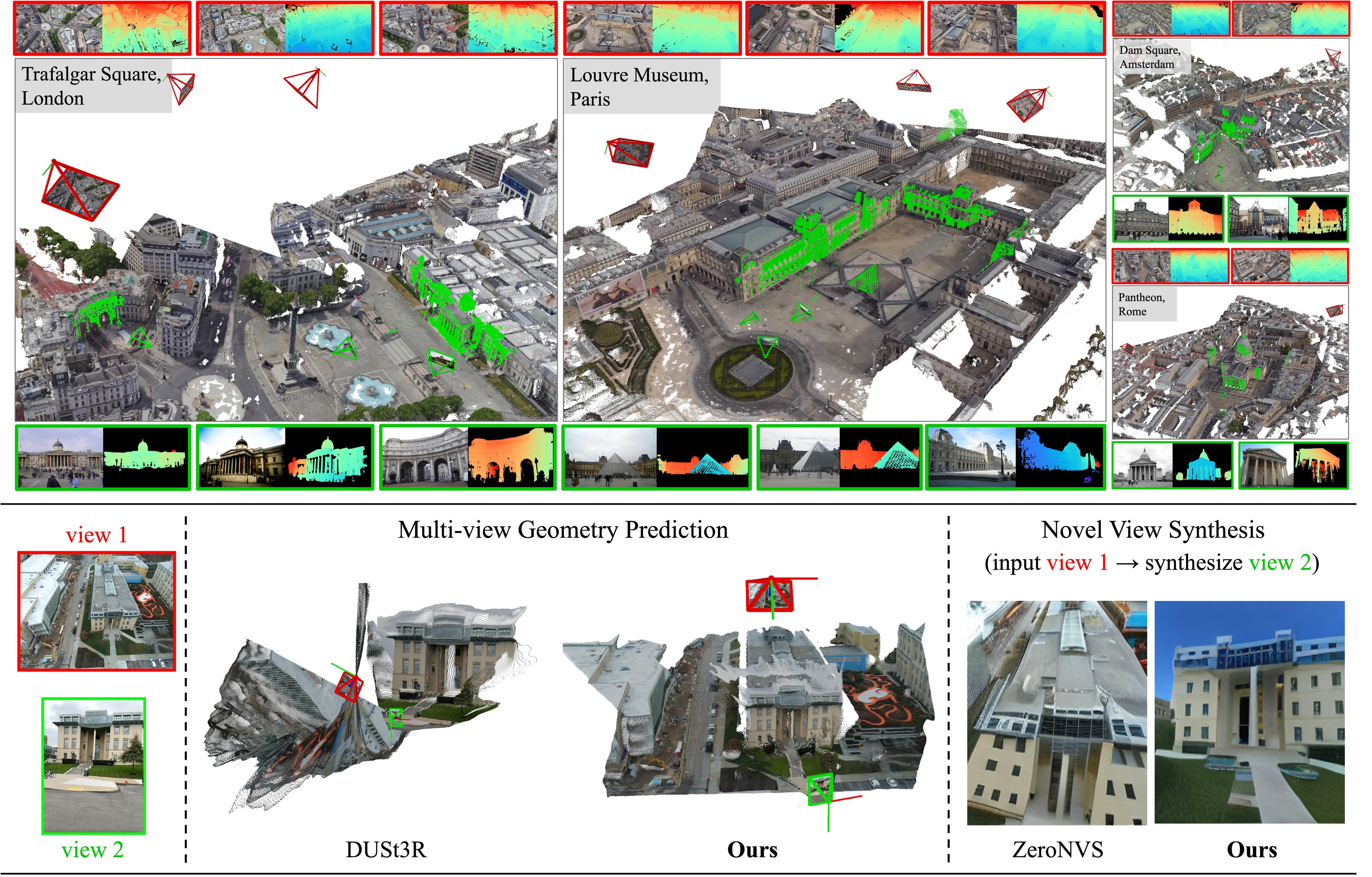}
    \vspace{-0.7cm}
    \captionof{figure}{\textbf{First row:} Examples of our generated cross-view (aerial-ground) geometry data, including co-registered \textcolor{red}{pseudo-synthetic (i.e., mesh-rendered) aerial} and \textcolor{mygreen}{real ground-level} images, with corresponding depth maps, point clouds, and camera intrinsics/extrinsics in a unified coordinate system, for a variety of scenes. \textbf{Second row:} Leveraging such data curated over 137 landmarks and 132K geo-registered images, we show significant improvements in learning-based methods on \textbf{real unseen} ground-aerial scenarios across two representative tasks: 1) multi-view geometry prediction using DUSt3R~\cite{dust3r_cvpr24} finetuned on our data, and 2) novel view synthesis from a single image conditioned on a target pose by fine-tuning ZeroNVS~\cite{zeronvs} that was originally trained on MegaScenes~\cite{tung2024megascenes}.}
    \label{fig:teaser}
\end{center}
}]

{\let\thefootnote\relax\footnote{{$^{\ast}$ denotes equal contribution/advising}}}

\begin{abstract}
We explore the task of geometric reconstruction of images captured from a mixture of ground and aerial views.
Current state-of-the-art learning-based approaches fail to handle the extreme viewpoint variation between aerial-ground image pairs.
Our hypothesis is that the lack of high-quality, co-registered aerial-ground datasets for training is a key reason for this failure. Such data is difficult to assemble precisely because it is difficult to reconstruct in a scalable way.
To overcome this challenge, we propose a scalable framework combining pseudo-synthetic renderings from 3D city-wide meshes (e.g., Google Earth) with real, ground-level crowd-sourced images (e.g., MegaDepth~\cite{megadepth}). The pseudo-synthetic data simulates a wide range of aerial viewpoints, while the real, crowd-sourced images help improve visual fidelity for ground-level images where mesh-based renderings lack sufficient detail, effectively bridging the domain gap between real images and pseudo-synthetic renderings.
Using this hybrid dataset, we fine-tune several state-of-the-art algorithms and achieve significant improvements on real-world, zero-shot aerial-ground tasks. 
For example, we observe that baseline DUSt3R~\cite{dust3r_cvpr24} localizes fewer than 5\% of aerial-ground pairs within 5 degrees of camera rotation error, while fine-tuning with our data raises accuracy to nearly 56\%, addressing a major failure point in handling large viewpoint changes.
Beyond camera estimation and scene reconstruction, our dataset also improves performance on downstream tasks like novel-view synthesis in challenging aerial-ground scenarios, demonstrating the practical value of our approach in real-world applications.

\end{abstract}
\vspace{-0.15in}
    
\section{Introduction}
\label{sec:intro}

The ability to register, reconstruct, or generally reason about multi-view images has been a cornerstone task in computer vision.  
While classical pipelines~\cite{colmap,schoenberger2016mvs} leveraged hand-designed features~\cite{sift,rublee11orb,bay2006surf} and matching mechanisms, their subsequent incarnations~\cite{pan2024glomap, wang2023vggsfm, brachmann2024acezero} have incorporated several learning-based components \eg learned features~\cite{superpoint,pixsfm} or learned ``doppelganger'' detectors~\cite{cai2023doppelgangers}. More recently, approaches have departed from this classical pipeline to directly learn multi-view tasks such as 2D correspondences \cite{sun2021loftr, superglue, lightglue}, camera estimation~\cite{lin2024relposepp,zhang2024raydiffusion, wang2023posediffusion}, pointmap prediction~\cite{dust3r_cvpr24, mast3r_arxiv24} and novel-view synthesis~\cite{tung2024megascenes,zeronvs, liu2023zero1to3} in an end-to-end manner. This shift towards learning-based components and approaches has led to impressive progress, particularly in challenging scenarios, e.g., sparsely sampled input, or varying illumination.

Much of this progress has been fueled by large crowd-sourced image collections like MegaDepth~\cite{megadepth}, which provides accurate 3D reconstructions built using structure-from-motion (SfM) from thousands of tourist-uploaded images at various landmarks. This 3D data offers valuable supervision for geometric tasks and has been transformative for multi-view learning algorithms.
However, because these images are primarily captured by tourists, they mostly cover ground-level viewpoints, and in some cases, aerial viewpoints, but rarely both.
As a result, methods like DUSt3R~\cite{dust3r_cvpr24} and MASt3R~\cite{mast3r_arxiv24}, though trained on diverse ``in-the-wild'' datasets including MegaDepth, struggle with large viewpoint changes between handheld (ground-level) and drone-mounted (aerial) views, as shown in Figure~\ref{fig:teaser}.
For example, in our evaluation, pre-trained DUSt3R achieves only a $\sim$5\% success rate in registering cameras (under $5\degree$ rotational error) from ground-aerial pairs.

Our central hypothesis is that this limitation stems from the lack of training data that contains co-registered ground-aerial image pairs. While independent ground and aerial camera poses are easy to obtain, merging them into a unified coordinate system often requires specialized sensors or manual effort, limiting scalability. 
To address this, we propose a flexible and scalable data generation framework leveraging geospatial platforms like Google Earth, which render 3D textured meshes of cities and landmarks -- providing another vast data source.
We refer to these mesh renderings as \textit{pseudo-synthetic} since they are rendered from 3D meshes of actual landmarks textured with real photos.
We construct pseudo-synthetic data by rendering aerial-ground viewpoints at varying altitudes from these textured meshes. While using these images alone show promising improvements, such mesh renderings from ground-level viewpoints are not as photorealistic, with a domain gap due to differences in lighting, texture, and other visual details compared to real images.
To mitigate this, we propose to co-register abundantly available real ground images (e.g., from MegaDepth~\cite{megadepth}) with the pseudo-synthetic images within the same coordinate frame. 
The pseudo-synthetic data, captured at varying altitudes, simulates a wide range of aerial viewpoints, while the real, crowd-sourced images help improve visual fidelity, especially for ground-level images where mesh renderings lack details. We call this hybrid dataset \textit{AerialMegaDepth}, and Figure~\ref{fig:teaser} shows some landmarks from our data, with 132,137 co-registered real and pseudo-synthetic images across 137 scenes.

 From this dataset, we generate over 1.5 million aerial-ground image pairs and fine-tune several state-of-the-art reconstruction algorithms, showing significant improvements on real-world mixed-altitude imagery (see Figure~\ref{fig:teaser} for examples).
Quantitatively, fine-tuning 3D prediction models like DUSt3R~\cite{dust3r_cvpr24} and MASt3R~\cite{mast3r_arxiv24} improves the camera registration success rate (with rotation error under 5\degree) from just $5\%$ to nearly $56\%$.
In addition, our dataset also improves novel-view synthesis in challenging aerial-ground scenarios (see Figure~\ref{fig:teaser}).
Finally, we emphasize that our framework is flexible and scalable, applicable not only to MegaDepth but also to other crowd-sourced datasets~\cite{Mapillary, BingStreetside, GoogleStreetView, tung2024megascenes} and  geospatial platforms~\cite{deschaud2021paris, lei2023assessing, Cesium}, making it possible to leverage a nearly unlimited source of data to learn aerial-ground 3D reconstruction.

\section{Related Works}
\label{sec:related}

\begin{figure*}[tbh]
    \centering
    \includegraphics[width=0.95\textwidth]{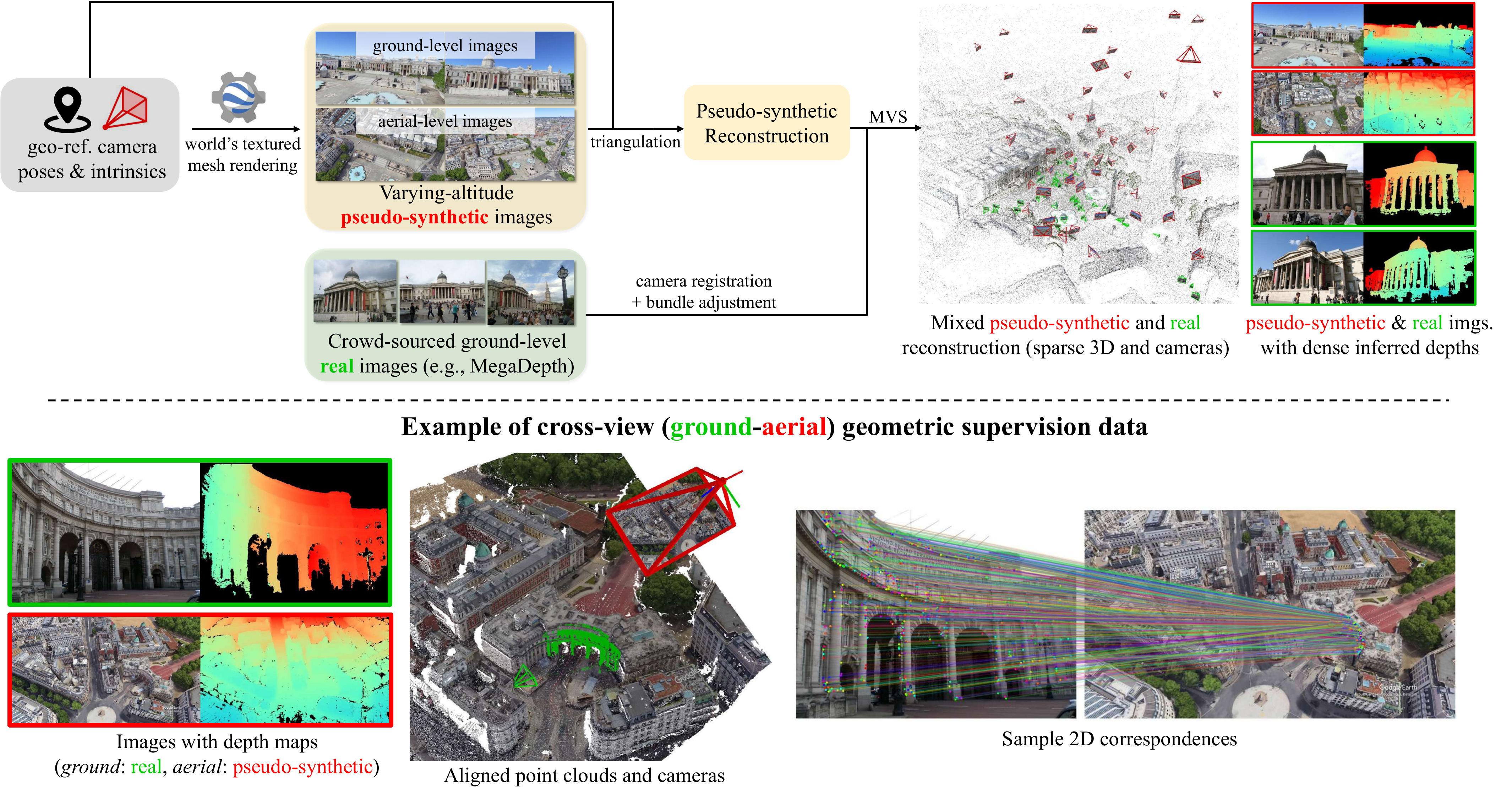}
    \vspace{-0.15in}
    \caption{\textbf{Overview of the data generation framework.} To address the challenges of ground-aerial camera registration and novel-view synthesis, we propose a flexible framework combining \textcolor{red}{pseudo-synthetic} renderings from 3D city-wide meshes (\eg Google Earth) with \textcolor{mygreen}{real}, ground-level images (\eg MegaDepth~\cite{megadepth}). The pseudo-synthetic data is captured at varying altitudes, while the real, crowd-sourced images help improve visual fidelity especially for ground-level images where mesh-based renderings lack detail. The pipeline generates pseudo-synthetic images from different altitudes, co-registers them with real images, and aligns ground-level images with aerial data for 3D reconstruction. This hybrid dataset of real and pseudo-synthetic images provides geometric supervision that helps improve performance on downstream tasks such as ground-aerial camera registration and novel view synthesis, particularly in ground-aerial settings.}
    \label{fig:data_generation}
    \vspace{-0.2in}
\end{figure*}

\noindent \textbf{Datasets for Multiview 3D and Geometry.} Several datasets released in the last decade have pushed the frontier of multiview 3D geometry. Internet-sourced datasets of landmarks like MegaDepth~\cite{megadepth} and IMC-PT~\cite{jin2021image} mostly have ground-level views. But still, these datasets, whose ground-truth was built using SfM~\cite{colmap}, have spurred monocular depth estimation~\cite{zoedepth, depthanything}, multiview stereo~\cite{dust3r_cvpr24, mast3r_arxiv24}, learned feature matching~\cite{superglue, sun2021loftr, roma, patch2pix}, learned pose regression~\cite{wang2023posediffusion, wang2023vggsfm, zhang2024raydiffusion}, etc. Datasets like BlendedMVS~\cite{blendedMVS} generate training images by blending rendered images from textured 3D mesh with input real images. However, scenes in these datasets are \textit{solely} captured from drones or ground, and not both simultaneously.
This is because capturing data with drastic viewpoint changes is challenging and requires specialized sensors or manual effort. Instead, our pseudo-synthetic data, paired with real ground images, anchors scenes with simultaneous aerial and ground views.

Datasets with 3D city meshes~\cite{deschaud2021paris, lei2023assessing} have also been proposed. 
In contrast, our dataset contains both real and pseudo-synthetic data of the same landmarks in a unified coordinate system. While purely synthetic data~\cite{li2023matrixcity, raistrick2023infinite, Dosovitskiy17, shah2018airsim, zhang2024drone} have been used for various 3D learning tasks, these datasets are not anchored to real scenes, making it difficult to bridge the sim-to-real gap. Methods using 3D meshes with image queries have been explored for other tasks like visual place recognition~\cite{Berton_ECCV_2024_MeshVPR, vallone2022danish} and localization~\cite{panek2022meshloc, panek2023visual}.

\noindent \textbf{Learning for Multiview 3D.}
Driven by large-scale supervision, learning-based methods have resulted in more robust feature matching~\cite{superglue, sun2021loftr, aspanformer22, lightglue, edstedt2023dkm, roma, edstedt2024dedode}, pose estimation~\cite{dust3r_cvpr24, zhang2024raydiffusion, zhang2022relpose, wang2023posediffusion} and direct regression of 3D pointmaps in a unified coordinate frame~\cite{dust3r_cvpr24, mast3r_arxiv24}. Likewise, generating novel views from an input view conditioned on camera poses~\cite{zhou2023sparsefusion, liu2023zero1to3, zeronvs, shi2023MVDream} is enabled by large datasets like MegaScenes~\cite{tung2024megascenes}. However, these methods struggle with extreme ground-aerial viewpoints due to limited supervision data. We show that our hybrid real and pseudo-synthetic data significantly improves performance of such learning-based methods on real mixed-altitude imagery.

\noindent \textbf{Aerial-ground Registration.} Prior work has addressed aerial-ground registration using specialized classical and learning-based methods~\cite{shan2014accurate, lin2015learning, 10131998, zhu2020leveraging}. Our work is orthogonal, as we propose a hybrid dataset to potentially improve these methods among other tasks. In the same vein, very few co-registered ground-aerial datasets exist. MAVREC~\cite{Dutta_2024_CVPR} and University1652~\cite{zheng2020university} provides no depth or pose information, and GrAco~\cite{graco} has only monochrome imagery and is small-scale.
We also distinguish our work from satellite-ground localization~\cite{sarlin2024snap, fervers2023uncertainty, shi2022beyond, shi2019spatial, workman2015wide} as satellite viewpoint is orthographic and distant while viewpoints in our data are closer to a drone-view sharing common (albeit small) field-of-view with the ground-view.

\section{Generating Aerial-Ground 3D Data}
\label{sec:method}

To address the scarcity of aerial-ground 3D data, we introduce an approach that combines renderings from 3D city-wide mesh models with real crowd-sourced images. The 3D meshes generate pseudo-synthetic renderings across varied altitudes and orientations, particularly aerial viewpoints, while real images complement this with ground-level captures, where pseudo-synthetic renderings often lack visual fidelity. An overview of our framework is in Figure~\ref{fig:data_generation}.

\subsection{Pseudo-synthetic data generation}

We chose Google Earth as our primary data source for its quality and landmark coverage, allowing us to render images from any viewpoint. Our framework, however, is compatible with any geo-referenced 3D textured meshes~\cite{Berton_ECCV_2024_MeshVPR, panek2023visual, Cesium}. 
These images are termed \textit{pseudo-synthetic} as they are renderings of 3D meshes textured with real photos.

\noindent \textbf{Automatically generating query viewpoints.}
Our goal is to render images that have sufficient visual overlap, both with each other and with the real images from MegaDepth~\cite{megadepth}.
We start with scenes from MegaDepth which contains SfM reconstructions for 196 landmarks, each with thousands of internet-sourced images. While the EXIF GPS tags in these images are not precise enough for accurate co-registration or alignment, we use them to roughly sample Google Earth’s rendering viewpoints, ensuring they correspond to the same building or landmark. This is achieved by geo-referencing the local 3D reconstruction into the global ECEF frame using a similarity transform computed from the noisy GPS data.
To ensure co-visibility among rendered images, we sample 200 points from the geo-referenced point cloud to serve as \textit{look-at} targets for generating query viewpoints.
This ensures that our rendered pseudo-synthetic images maintain sufficient visual overlap with each other and also with MegaDepth’s real images.

\noindent \textbf{Generating pseudo-synthetic 3D reconstruction.}
While users can specify and render images from any location and orientation, Google Earth unfortunately does not provide direct access to the underlying 3D mesh. Therefore, to recover scene geometry from the rendered pseudo-synthetic images (for which camera extrinsics and intrinsics are known), we extract keypoints~\cite{superpoint} and match features~\cite{lightglue} between the pseudo-synthetic rendered images to triangulate the 3D point cloud. 
With this process, we generated data from 137 sites or landmarks, each with 600 images taken from varying elevations, ranging from 1 meter to 350 meters, resulting in a total of 82,220 pseudo-synthetic images.
We refer to this as \textit{pseudo-synthetic reconstruction}.

\subsection{Co-registering real crowd-sourced images}

\begin{figure}[t!] \centering
    \includegraphics[width=\columnwidth]{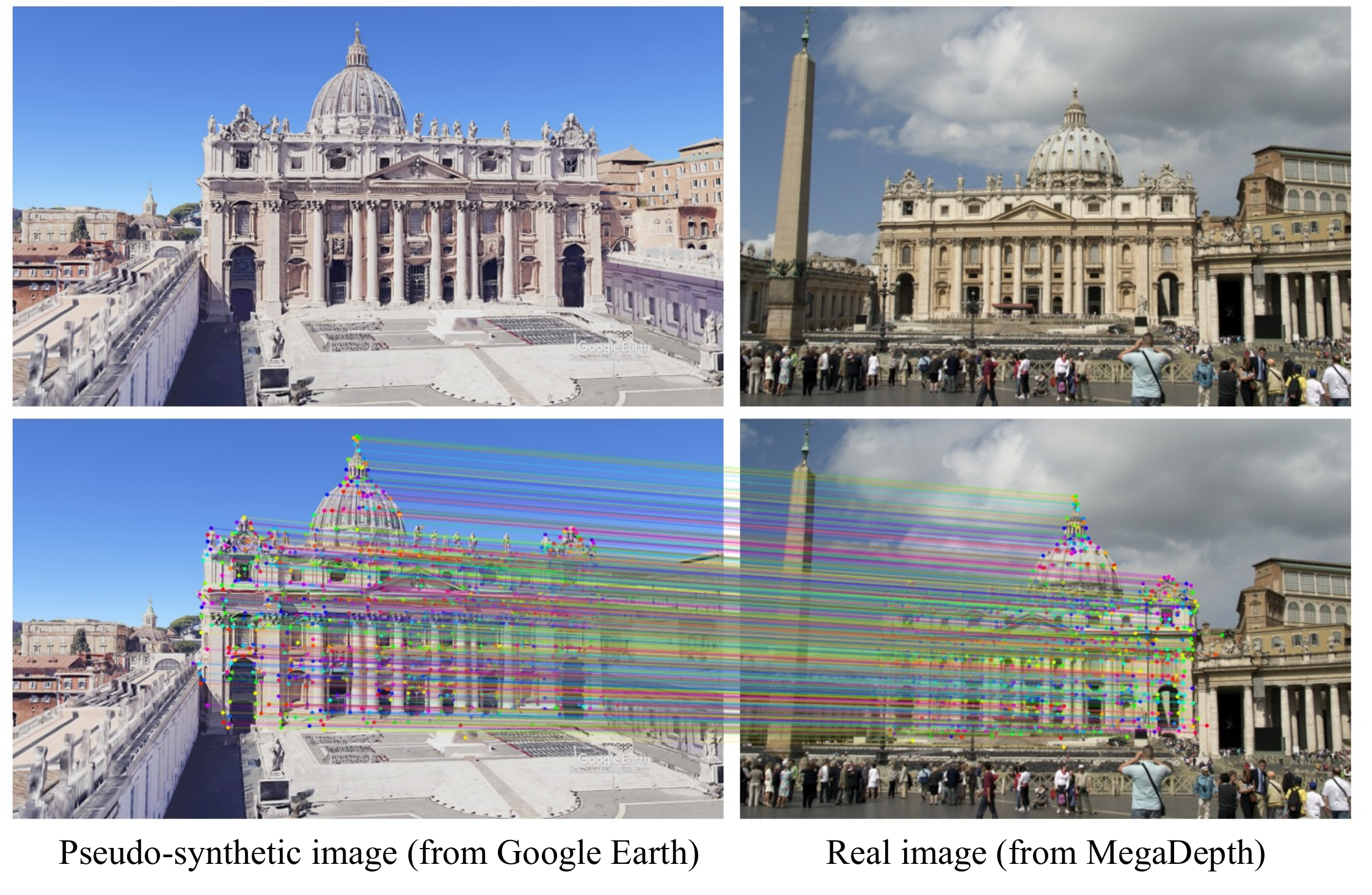}
    \vspace{-0.3in}
    \caption{\textbf{Feature matching between real and pseudo-synthetic images.} The pseudo-synthetic rendering has a noticeable domain gap compared to the real MegaDepth image (e.g., no transients, simplistic lighting) but still enables reliable feature matching~\cite{superglue} to register real images into the pseudo-synthetic reconstruction.}
    \label{fig:pseudo_real_matches}
    \vspace{-0.2in}
\end{figure}

While pseudo-synthetic images from 3D textured meshes can directly improve geometry prediction tasks -- as demonstrated in Section~\ref{sec:experiments} -- we make two key observations that help further improve the downstream performance.
First, since these meshes are typically textured from aerial images, they tend to appear realistic from elevated, drone-like viewpoints but often lack visual fidelity at ground level, where low-viewing angles introduce artifacts on building facades~\cite{panek2023visual}.
A domain gap also exists between ground-level pseudo-synthetic and real images, such as the absence of transients and simplistic lighting model, which may limit generalization to real-world data.
Second, despite these limitations, mesh-based visual localization methods~\cite{panek2023visual} show that real images can still be registered accurately to pseudo-synthetic reconstructions with state-of-the-art feature matching~\cite{superglue, lightglue, sun2021loftr} as shown in Figure~\ref{fig:pseudo_real_matches}.
To combine the benefits of both real and pseudo-synthetic images, we thus register real, ground-level images into the pseudo-synthetic reconstructions, resulting in aligned real (ground) MegaDepth and pseudo-synthetic (aerial) images. 

\begin{figure}[tp]
    \centering
\includegraphics[width=\columnwidth]{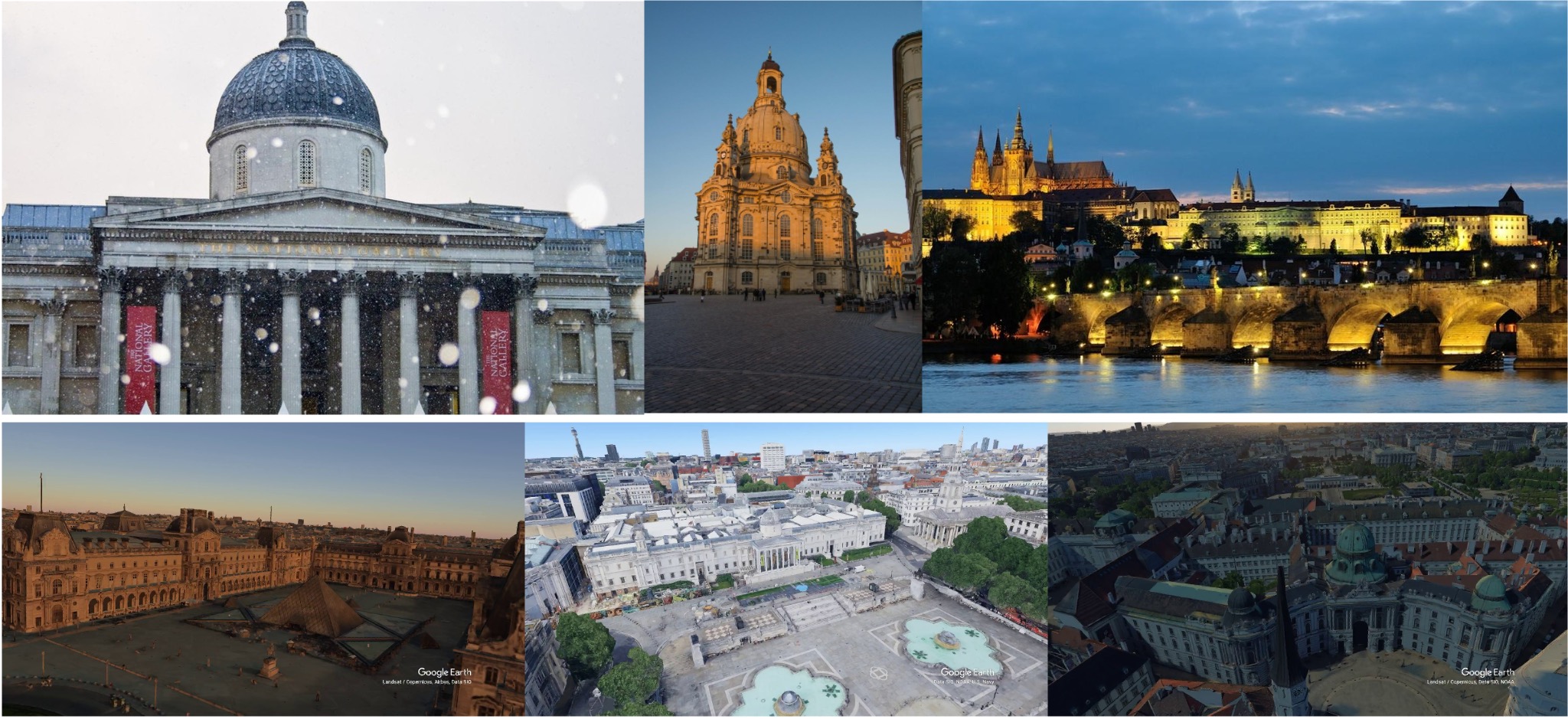}
    \vspace{-0.2in}
    \caption{AerialMegaDepth data (top: MegaDepth, bottom: Google Earth) features \textbf{diverse viewpoints \& lighting conditions}.}
    \label{fig:training_lv}
   \vspace{-0.1in}
\end{figure}

Specifically, we follow standard visual localization pipeline~\cite{hloc} by first retrieving top-$k$ most similar pseudo-synthetic images for each real MegaDepth query image~\cite{arandjelovic2016netvlad}. 2D correspondences~\cite{superpoint, lightglue} are lifted to 2D-3D matches using the pseudo-synthetic 3D points, and each query image's 6-DoF pose is estimated with a RANSAC-based PnP solver~\cite{lepetit2009ep}. We  refine the alignment by optimizing the localized MegaDepth images while keeping pseudo-synthetic cameras fixed.
Using COLMAP's MVS~\cite{schoenberger2016mvs}, we generate semi-dense depth maps for supervision. In total, we register 49,937 MegaDepth images with 82,200 pseudo-synthetic images across 137 scenes, forming \textit{AerialMegaDepth} -- a hybrid dataset of 132,137 images with diverse viewpoints and lighting variations, as shown in Figure~\ref{fig:training_lv}.

\begin{figure*}[tp]
    \centering
    \includegraphics[width=0.95\textwidth]{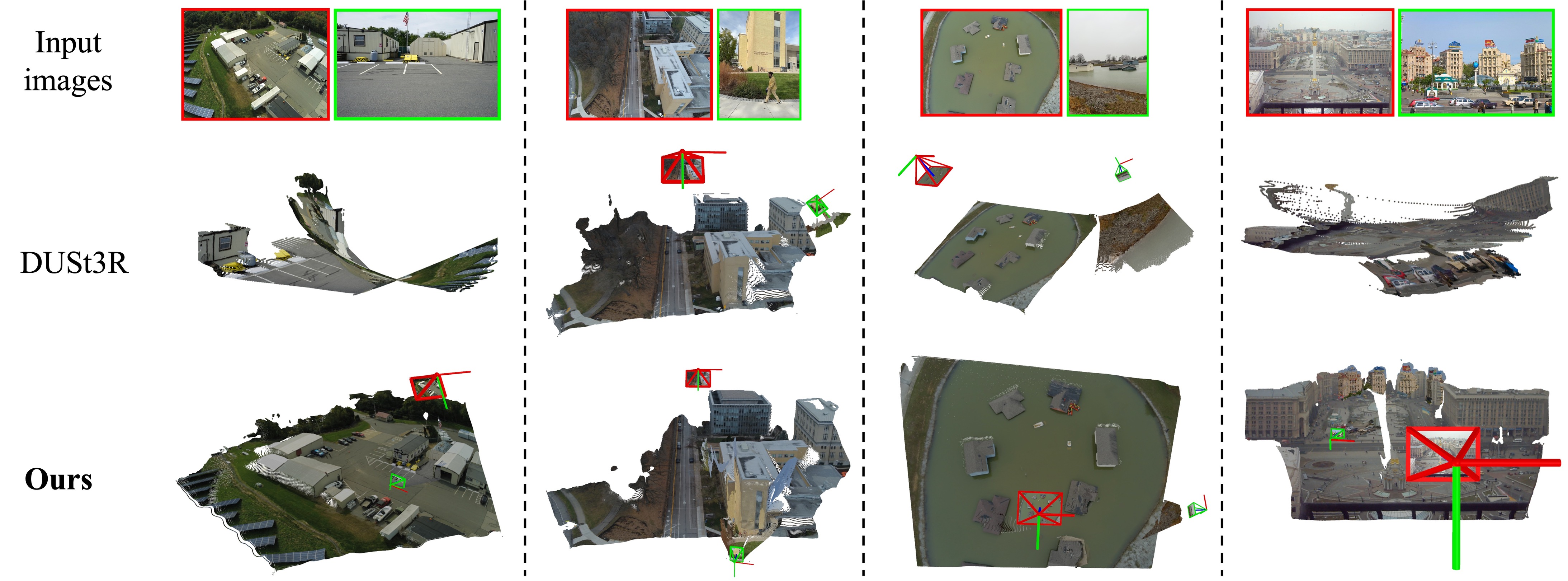}
    \vspace{-0.1in}
    \caption{\textbf{Zero-shot ground-aerial camera and geometry prediction results.} Given two input images, one aerial and one ground, we compare the performance of the baseline DUSt3R~\cite{dust3r_cvpr24} with the model fine-tuned on our varying-altitude data. The results demonstrate significant improvements over the baseline in \textbf{unseen, challenging} ground-aerial scenarios, showing the effectiveness of fine-tuning DUSt3R~\cite{dust3r_cvpr24} with our data. Additionally, the last column presents qualitative results on a challenging ground-aerial pair from the WxBS~\cite{wxbs} dataset, which involves significant viewpoint change. }
    \label{fig:ground-aerial-pairwise}
   \vspace{-0.1in}
\end{figure*}

\section{Learning Aerial-Ground 3D Reconstruction}
We explore our data's impact on  supervised learning for multi-view 3D reconstruction and novel view synthesis. %

\noindent \textbf{Selection of image pairs as supervision data.}
Our objective is to select image pairs that offer adequate overlap for effective supervision, particularly for ground-to-aerial pairs with altitude differences, ensuring the overlap is neither too high (making the task too easy) nor too low (making it too difficult).
To achieve this, we compute an $N \times N$ covisibility matrix $\mathcal{C}$ for each scene, where each element $\mathcal{C}[i, j] \in [0, 1]$ represents the percentage of points in image $i$ that are visible in image $j$. 
For ground-to-aerial settings, we prioritize pairs with significant viewpoint differences, meaning the covisibility is asymmetrical: we select pairs $(i, j)$ such that $\mathcal{C}[i, j]$ is small and $\mathcal{C}[j, i]$ is large, or vice-versa.
We quantify this difference with a score $s=\frac{\mathrm{AM}}{\mathrm{HM}}$, where $\mathrm{AM}$ is the arithmetic mean and $\mathrm{HM}$ is the harmonic mean of $\mathcal{C}[i, j]$ and $\mathcal{C}[j, i]$.
A high score indicates pairs with a large  viewpoint difference, ideal for challenging cross-view tasks, aligning with our goal of prioritizing ground-to-aerial pairs.
Using this approach, we generate a varying-altitude dataset of 1.5M image pairs, each comes with camera intrinsics, camera poses, and depthmaps to be used as supervision for learning geometric tasks.

\noindent \textbf{Multi-view Pose and Geometry Estimation.} We consider the problem of estimating intrinsic, extrinsic camera parameters, and 3D scene from a set of $N$ unconstrained images. Following the architecture of DUSt3R~\cite{dust3r_cvpr24}, we regress pointmaps for image pairs, where each pixel is associated with a 3D point in the coordinate system of the first frame. This allows us to compute the per-camera focal length (under a pinhole model with centered principal point) and the 6-DoF relative pose $\mathbf{T} = [\mathbf{R} | \mathbf{t}]$ for each image pair using PnP~\cite{lepetit2009ep}.
For $N > 2$ images, DUSt3R's global alignment (GA) step combines pointmaps across all images, optimizing a dense pairwise graph in 3D to align pointmaps within a global coordinate frame.
We initialize with a DUSt3R checkpoint trained on millions of image pairs from eight datasets~\cite{co3d, scannet++, arkitscenes, megadepth, blendedMVS, Savva_2019_ICCV, Sun_2020_CVPR, MIFDB16} and fine-tune on our data, resulting in substantial improvements for ground-aerial camera registration.
We also observe similar improvements by fine-tuning MASt3R~\cite{mast3r_arxiv24} and using it as a front-end to provide 2D-2D correspondences, which are then fed into COLMAP for bundle adjustment (similar to the approach in MASt3R-SfM~\cite{duisterhof2024mast3r}).

\noindent \textbf{Novel View Synthesis.} For single-image novel view synthesis (NVS), our goal is to synthesize a plausible target ground view from a reference aerial image. Tung et al.~\cite{tung2024megascenes} fine-tuned ZeroNVS~\cite{zeronvs} on MegaScenes with over 2M image pairs from 32K scenes, achieving significant improvements on scene-level view synthesis compared to object-centric settings~\cite{dtu, co3d}. 
By further fine-tuning ZeroNVS on our dataset with varying altitudes, we achieve significant improvements in aerial-to-ground view synthesis.

\section{Experiments}
\label{sec:experiments}

\begin{figure}[tp]
    \centering
    \includegraphics[width=\columnwidth]{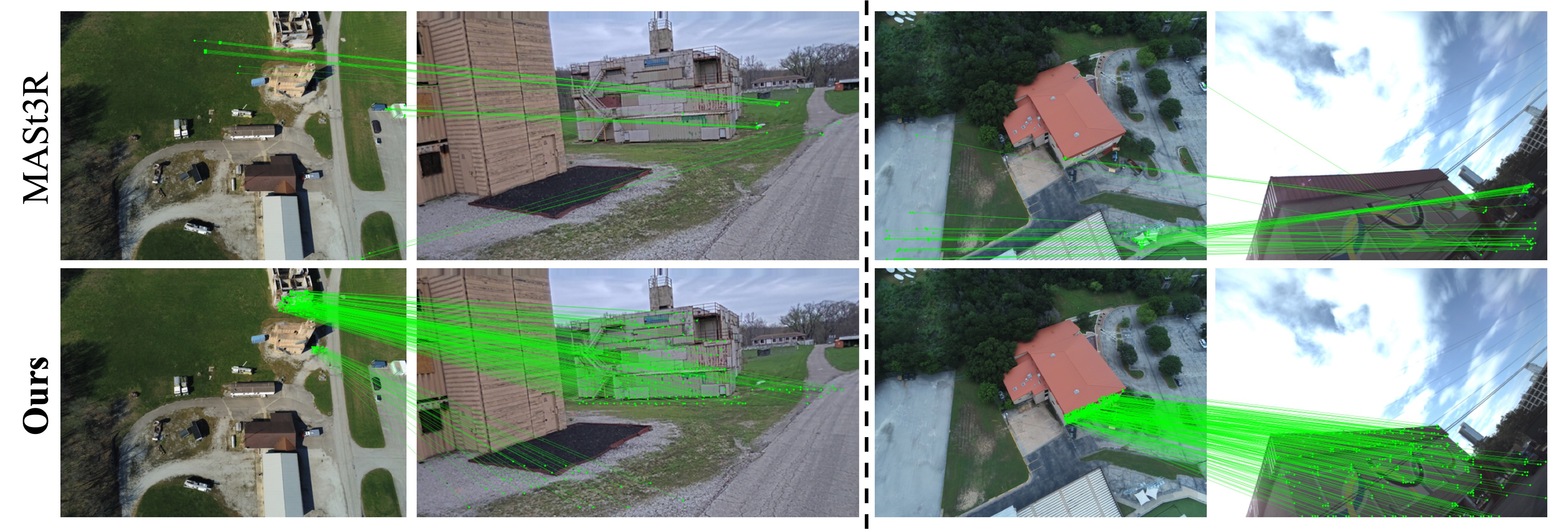}
    \vspace{-0.25in}
    \caption{\textbf{Challenging ground-aerial feature matching.} Fine-tuned MASt3R~\cite{mast3r_arxiv24} achieves accurate and robust feature matching across ground-aerial pairs with extreme viewpoint changes (correspondences extracted via reciprocal nearest neighbor from MASt3R's local feature maps). This highlights the effectiveness of our AerialMegaDepth data in improving matching performance.}
    \label{fig:mast3r_matches}
   \vspace{-0.2in}
\end{figure}

\definecolor{first}{rgb}{1.0, .83, 0.3}
\definecolor{second}{rgb}{1.0, 0.93, 0.7}
\def \first {\cellcolor{first}}
\def \second {\cellcolor{second}}

\begin{table*}[h]
\centering
\resizebox{0.95\textwidth}{!}{%
\begin{tabular}{l|ccc|ccc|ccc}
\toprule
\multicolumn{1}{c|}{\multirow{2}{*}{Method}} & \multicolumn{3}{c|}{\emph{Camera Rotation Accuracy}} & \multicolumn{3}{c|}{\emph{Camera Translation Accuracy}} & \multicolumn{3}{c}{\emph{3D Pointmap Accuracy}} \\ 
\cline{2-10} 
\multicolumn{1}{c|}{}                        & $\RRA$@5\degree              & $\RRA$@10\degree             & $\RRA$@15\degree            & $\RTA$@5\degree             & $\RTA$@10\degree            & $\RTA$@15\degree           & $\delta$@0.5\text{m}           & $\delta$@1\text{m}          & $\delta$@2\text{m}          \\ \midrule

LoFTR~\cite{sun2021loftr}                                & 0.92 & 1.83 & 2.45 & 0.92 & 1.53 & 2.14 & - & - & - \\ 
SP+SG~\cite{superglue}                               & 8.56 & 10.09 & 12.23 & 7.65 & 9.79 & 11.31 & - & - & - \\
MASt3R~\cite{mast3r_arxiv24} (released)                                       & 3.36 & 3.36 & 4.59 & 2.45 & 3.36 & 4.28 & - & - & - \\
MASt3R + MatrixCity                                & 19.78 & 30.88 & 38.17 & 10.67 & 25.72 & 29.43 &  - & - & - \\ 
\textbf{MASt3R + PSynth (Ours)}                                & \second 26.49 & \second 43.71 & \second 47.62 & \second 25.25 & \second 40.32 & \second 49.34 & - & - & - \\
\textbf{MASt3R + Hybrid (Ours)}                                & \first \textbf{49.54} & \first \textbf{66.36} & \first \textbf{72.48} & \first \textbf{42.51} & \first \textbf{63.30} & \first \textbf{69.11} & - & - & - \\
\midrule
DUSt3R~\cite{dust3r_cvpr24} (released)                                       & 5.20 & 7.95 & 9.48 & 2.75 & 5.81 & 9.17 & 29.02 & 42.16 & 43.79 \\
DUSt3R + MatrixCity                                & 17.85 & 37.28 & 42.80 & 11.33 & 25.24 & 33.24 & 31.43 & 47.13 & 57.02 \\ 

\textbf{DUSt3R + PSynth (Ours)}                                & \second 31.28 & \second 47.63 & \second 51.61 & \second 28.78 & \second 45.66 & \second 51.47 & \second 32.77 & \second 53.42 & \second 61.45 \\

\textbf{DUSt3R + Hybrid (Ours)}                                & \first \textbf{55.96} & \first \textbf{71.25} & \first \textbf{76.15} & \first \textbf{46.48} & \first \textbf{68.20} & \first \textbf{72.78} & \first \textbf{38.24} & \first \textbf{62.33} & \first \textbf{74.52} \\   
                                                  
 \bottomrule 
\end{tabular}
}
\vspace{-0.1in}
\caption{\textbf{Finetuning with our data significantly improves pairwise camera pose estimation in the \textit{ground-aerial} setting.} Baselines, including learned 2D correspondence matching (SP+SG~\cite{superpoint, superglue}, LoFTR~\cite{sun2021loftr}, MASt3R~\cite{mast3r_arxiv24}) and 3D pointmap-based regression (DUSt3R~\cite{dust3r_cvpr24}), struggle in this setting. For instance, DUSt3R localizes fewer than 5\% of pairs within 5\degree rotation error ($\RRA@5\degree$). Finetuning on MatrixCity improves performance, but using pseudo-synthetic ground-aerial pairs (DUSt3R + PSynth) boosts accuracy to 31\%, and adding real ground data (DUSt3R + Hybrid) further increases it to 55\%. This also significantly improves 3D pointmap accuracy. The first half of the table shows methods that predict 2D matches, with ground-truth intrinsics used to compute the relative poses.}
\label{tab:pairwise_results}
\vspace{-0.1in}
\end{table*}

\subsection{Multiview Pose Estimation and Reconstruction}

\noindent \textbf{Datasets.} We fine-tune the baseline models~\cite{dust3r_cvpr24, mast3r_arxiv24} on our data using pairwise 3D pointmaps as supervision.
For evaluation, we focus on ground-aerial settings and include data from ULTRRA Challenge~\cite{ultrra}, which consists of images captured by ground-level cameras and drones, all calibrated using SfM constrained by RTK-corrected GPS coordinates for cm-level accuracy.
Additionally, we use data from ACC-NVS1~\cite{sugg2025accenturenvs1novelviewsynthesis} which captures various urban sites with ground-truth poses obtained via GNSS/IMU systems corrected by a stationary RTK base station.
Overall, the evaluation data includes six sites with over 5,000 calibrated ground-aerial images.

\noindent \textbf{Evaluation Metrics.} Following~\cite{wang2023posediffusion, jin2021image}, we evaluate camera pose using \emph{Relative Rotation Accuracy} ($\RRA$) and \emph{Relative Translation Accuracy} ($\RTA$). $\RRA$ measures the angular difference between the predicted and ground-truth relative rotations, and $\RTA$ calculates the angular difference between the predicted and ground-truth translation vectors.
We report $\RTA@\tau / \RRA@\tau$, i.e., the percentage of camera pairs with $\RTA / \RRA$ below a threshold $\tau$. 
We also evaluate reconstruction accuracy by aligning the predicted pointmap from DUSt3R with ground-truth from MVS~\cite{schoenberger2016mvs} using a RANSAC-based optimal similarity transform. We report $\delta$ @ [0.5m, 1m, 2m], representing the percentage of points with errors within 0.5m, 1m, and 2m, respectively.

\begin{figure*}[t]
    \centering
    \includegraphics[width=0.95\textwidth]{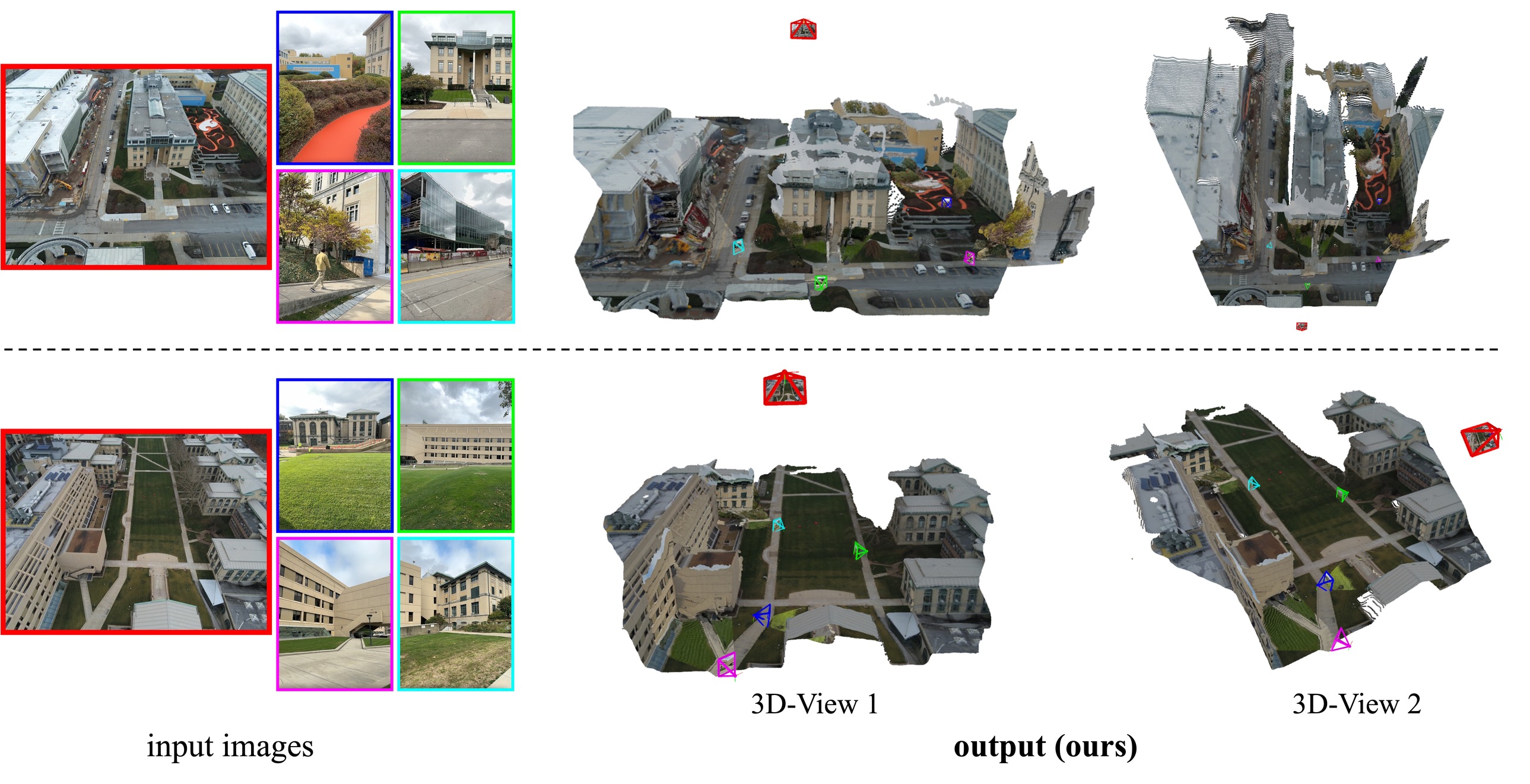}
    \vspace{-0.18in}
    \caption{\textbf{3D reconstruction from one aerial and four ground images with \emph{virtually no overlap}.} We use the global alignment process of DUSt3R~\cite{dust3r_cvpr24} to merge pointmaps predictions. Despite the lack of overlap among the ground images, we find that incorporating a reference aerial image can effectively serve as a ``map'', significantly improving pose estimation accuracy when fine-tuned on our cross-view data.}
    \label{fig:1aerial-Nground}
    \vspace{-0.1in}
\end{figure*}

\begin{table*}[t] \centering
\begin{minipage}[t]{0.49\linewidth}
    \resizebox{\textwidth}{!}{
    \large
    \begin{tabular}{lccccc}
\toprule
\# of ground images                        & 2 & 3 & 4 & 5 & 6 \\ \midrule
\textbf{no aerial image (i.e., ground only)} & \multicolumn{1}{l}{} & \multicolumn{1}{l}{} & \multicolumn{1}{l}{} & \multicolumn{1}{l}{} & \multicolumn{1}{l}{} \\
\multicolumn{1}{l}{DUSt3R-GA (released)}     & 12.20  & 32.21  & 38.31 & 43.98 & 47.98  \\ \midrule
\textbf{one aerial image} & \multicolumn{1}{l}{} & \multicolumn{1}{l}{} & \multicolumn{1}{l}{} & \multicolumn{1}{l}{} & \multicolumn{1}{l}{} \\
\multicolumn{1}{l}{DUSt3R-GA (released)}     & 14.63  & 33.02  & 37.50 & 43.73 & 48.47  \\
\multicolumn{1}{l}{DUSt3R-GA + PSynth (Ours)} & \second 29.27  & \second 44.72  & \second 48.78 & \second 55.85 & \second 55.45  \\
\multicolumn{1}{l}{DUSt3R-GA + Hybrid (Ours)} & \first \textbf{56.10}  & \first \textbf{55.28}  & \first \textbf{57.72} & \first \textbf{59.27} & \first \textbf{60.65} \\ 
\midrule
\multicolumn{1}{l}{MASt3R-SfM (released)} & 9.03 & 31.69 & 40.02 & 49.88 & 59.34 \\ 
\multicolumn{1}{l}{MASt3R-SfM + Psynth (Ours)} & \second 23.10 & \second 39.53 & \second 48.12 & \second 59.13 & \second 64.76 \\ 
\multicolumn{1}{l}{MASt3R-SfM + Hybrid (Ours)} & \first \textbf{51.07} & \first \textbf{52.21} & \first \textbf{61.31} & \first \textbf{63.92} & \first \textbf{67.45} \\ 
\bottomrule
\end{tabular}
    }
\end{minipage}\hfill
    \begin{minipage}[t]{0.49\linewidth}
    \resizebox{\textwidth}{!}{
    \large
    \begin{tabular}{lccccc}
\toprule
\# of ground images                        & 2 & 3 & 4 & 5 & 6 \\ \midrule
\textbf{no aerial image (i.e., ground only)} & \multicolumn{1}{l}{} & \multicolumn{1}{l}{} & \multicolumn{1}{l}{} & \multicolumn{1}{l}{} & \multicolumn{1}{l}{} \\
\multicolumn{1}{l}{DUSt3R-GA (released)}     & 9.76  & 27.96  & 31.40 & 40.80 & 43.13  \\ \midrule
\textbf{one aerial image} & \multicolumn{1}{l}{} & \multicolumn{1}{l}{} & \multicolumn{1}{l}{} & \multicolumn{1}{l}{} & \multicolumn{1}{l}{} \\
\multicolumn{1}{l}{DUSt3R-GA (released)}     & 9.76  & 27.15  & 31.40 & 41.78 & 43.62  \\
\multicolumn{1}{l}{DUSt3R-GA + PSynth (Ours)} & \second 31.27  & \second 43.09  & \second 46.82 & \second 55.72 & \second 56.10  \\
\multicolumn{1}{l}{DUSt3R-GA + Hybrid (Ours)} & \first \textbf{51.29}  & \first \textbf{52.85}  & \first \textbf{54.07} & \first \textbf{55.61} & \first \textbf{57.72} \\ 
\midrule
\multicolumn{1}{l}{MASt3R-SfM (released)} & 9.28 & 23.91 & 29.01 & 46.91 & 51.45 \\ 
\multicolumn{1}{l}{MASt3R-SfM + Psynth (Ours)} & \second 25.80 & \second 41.09 & \second 44.52 & \second 58.79 & \second 61.22 \\ 
\multicolumn{1}{l}{MASt3R-SfM + Hybrid (Ours)} & \first \textbf{48.84} & \first \textbf{49.71} & \first \textbf{57.89} & \first \textbf{60.98} & \first \textbf{62.41} \\ 
\bottomrule
\end{tabular}
    }
\end{minipage}

\vspace{-0.08in}
\caption{\textbf{Including a single aerial image with $N$ ground images notably improves pose estimation of the ground images}, as shown in Ground Cameras Rotation Accuracy @ $15\degree$ ($\RRA@15\degree$) (left) and Translation Accuracy @ $15\degree$ ($\RTA@15\degree$) (right). Using DUSt3R’s global optimization~\cite{dust3r_cvpr24}, Row 1 shows results for ground-only input images, while the rest includes an aerial image as input. Although pose estimation improves with more ground images, adding even one aerial reference image significantly boosts accuracy, especially when ground images have minimal overlap (e.g., $N \leq 3$) as this aerial view helps align the ground images within a shared coordinate frame.}
\label{tab:1aerial_Nground}
\vspace{-0.15in}
\end{table*}

\noindent \textbf{Two-view pose and geometry estimation.} We evaluate the impact of our data on ground-aerial registration in the two-view case (one aerial and one ground image).
For DUSt3R~\cite{dust3r_cvpr24}, relative pose is computed from the predicted 2D-3D matches using PnP~\cite{lepetit2009ep}.
We also present baselines of 2D correspondence matching with SuperPoint~\cite{superpoint} + SuperGlue~\cite{superglue} (SP+SG), semi-dense matching with LoFTR~\cite{sun2021loftr}, and MASt3R~\cite{mast3r_arxiv24}, where for MASt3R, 2D correspondences are extracted via reciprocal nearest neighbor from its dense local feature maps. For these methods, we compute relative pose assuming known ground-truth intrinsics, using essential matrix estimated from 2D matches.

Table~\ref{tab:pairwise_results} shows that all baseline methods, including SP+SG, LoFTR, DUSt3R and MASt3R, struggle with ground-aerial pairs. For example, baseline DUSt3R only recovers $5.20\%$ of the total number of image pairs with good accuracy ($\RRA @ 5\degree$).
While fine-tuning on synthetic data like MatrixCity~\cite{li2023matrixcity} notably improves baseline methods, fine-tuning on pseudo-synthetic renderings (with more realistic textures, containing varying-altitude mesh renderings from Google Earth, denoted as DUSt3R/MASt3R + PSynth) is more effective, increasing accuracy to $31.28\%$ at $\RRA@5\degree$. 
But the largest improvement comes from training on hybrid data (that aligns pseudo-synthetic mesh renderings to real-world images for pair construction), denoted as DUSt3R/MASt3R + Hybrid, bringing the performance to more than $55\%$. This demonstrates that our novel framework of hybrid real and pseudo-synthetic data significantly improves ground-aerial camera registration.
We note that while our primary evaluation focuses on the ground-aerial setting, additional results in the Supplementary show that DUSt3R and MASt3R still perform well on similar-viewpoint pairs (e.g., ground-ground and aerial-aerial) even after fine-tuning with our varying-altitude data.

In addition to pose accuracy, we also observe substantial improvements in 3D geometry prediction, particularly in 3D pointmap accuracy from DUSt3R. As shown in Table~\ref{tab:pairwise_results}, finetuning on our combined dataset (DUSt3R + Hybrid) improves the percentage of 3D points within a 1-meter error by $20\%$ compared to the baseline DUSt3R.
Since we align predicted and ground-truth pointmaps using a RANSAC-based similarity transform, baseline DUSt3R model could still achieve reasonable geometry accuracy by producing good depth estimates for at least one of the views, \textit{even if they struggle to register them together accurately} (as shown in pose accuracy).
By finetuning on our data, we see substantial  improvements in both 3D pointmap accuracy and ground-aerial pose registration, highlighting the impact on both pose and geometry prediction.
We show qualitative results in Fig.~\ref{fig:ground-aerial-pairwise} and Fig.~\ref{fig:mast3r_matches} and encourage readers to check the website for additional results.
We emphasize that this is zero-shot performance on unseen data, as there is no overlap between our training and evaluation scenes.

\begin{figure*}[tp!]
    \centering
    \includegraphics[width=0.8\textwidth]{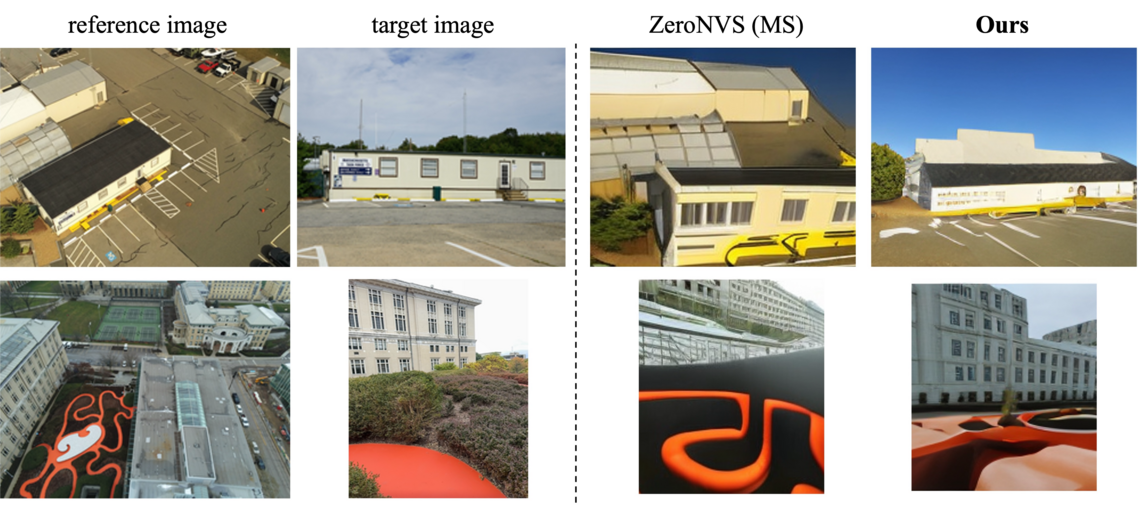}
    \vspace{-0.1in}
    \caption{\textbf{Results of extreme viewpoint change in novel-view synthesis} with ZeroNVS~\cite{zeronvs} finetuned on  MegaScenes~\cite{tung2024megascenes} (ZeroNVS MS) \& additionally finetuned on our data. Though by no means perfect, note the big improvement in visual quality and viewpoint accuracy.}    
    \label{fig:nvs_results}
    \vspace{-0.15in}
\end{figure*}

\noindent \textbf{Co-registering ground images with aerial context.} 
We evaluate multi-view ground-aerial camera registration, where one aerial image is matched with $N$ ground images. 
To perform multiview pose estimation, we explore two approaches: 1) the global alignment (GA) process of DUSt3R~\cite{dust3r_cvpr24} to merge all predicted pairwise pointmaps in the same coordinate frame, and 2) the MASt3R-SfM~\cite{duisterhof2024mast3r} approach, which uses MASt3R as a front-end to extract 2D correspondences followed by COLMAP~\cite{colmap} bundle adjustment.
In Table~\ref{tab:1aerial_Nground}, we report accuracy for ground images only.
Perhaps unsurprisingly, when ground images are sparse ($N \leq 3$), they often lack visual overlap, making pose estimation particularly challenging.
In this case, we observe a significant improvement in performance (from $14.63\%$ to $56.10\%$ for $\RRA@15\degree$ with $N=2$) when using the model finetuned with our data.
The key reason for this improvement is, even with almost no overlap among ground images, a single aerial image can serve as an ``overhead map'', helping to ``stitch'' or align the ground images into a common frame.
An example is shown in Figure~\ref{fig:1aerial-Nground} demonstrating the effectiveness of our approach in this scenario.

\subsection{Novel View Synthesis}

\noindent \textbf{Datasets.} 
We focus on the challenging aerial-to-ground synthesis task. 
We combine our dataset with MegaScenes~\cite{tung2024megascenes}, using a 3:1 ratio during finetuning to help prevent overfitting. 
For evaluation, we use both real-world aerial-ground pairs as well as pseudo-synthetic data from Google Earth that includes a wide range of images captured at varying altitudes, allowing us to assess the model's ability to synthesize views across diverse ground-aerial setups.

\noindent \textbf{Finetuning details.} We follow ZeroNVS~\cite{zeronvs} for novel view synthesis, which takes the extrinsic matrix and field-of-view as inputs to generate novel views at the target pose. The translation vector is scaled based on the 20th depth quantile of the reference image (during evaluation we used MVS depth). Starting from ZeroNVS trained on MegaScenes, we fine-tune the model on our dataset, significantly improving novel views in ground-aerial contexts.

\noindent \textbf{Evaluation Metrics.} We evaluate view synthesis quality using standard image reconstruction metrics, including LPIPS~\cite{zhang2018perceptual}, PSNR, and SSIM~\cite{wang2004image}. Additionally, we also include the DreamSim~\cite{fu2023dreamsim} score, which aligns more closely with human perceptual judgments.
\begin{table}[t!]
\centering
\resizebox{\columnwidth}{!}{%
\begin{tabular}{lcccc}
\toprule
NVS               & DreamSim$~\downarrow$             & LPIPS$~\downarrow$                & PSNR$~\uparrow$                 & SSIM$~\uparrow$                 \\ \midrule
\textbf{pseudo-synth. images} & \multicolumn{1}{l}{} & \multicolumn{1}{l}{} & \multicolumn{1}{l}{} & \multicolumn{1}{l}{} \\
ZeroNVS (MS)         & 0.448                & 0.413                & 10.847                & 0.416                \\
ZeroNVS (Ours)       & \textbf{0.377}                & \textbf{0.359}                & \textbf{12.381}                & \textbf{0.484}                \\ \midrule
\textbf{real images}          & \multicolumn{1}{l}{} & \multicolumn{1}{l}{} & \multicolumn{1}{l}{} & \multicolumn{1}{l}{} \\
ZeroNVS (MS)         & 0.550                & 0.639                & 7.478                & 0.183                \\
ZeroNVS (Ours)       & \textbf{0.442}                & \textbf{0.580}                & \textbf{8.220}                & \textbf{0.218}                \\ \bottomrule
\end{tabular}
}
\vspace{-0.1in}
\caption{\textbf{Quantitative results for aerial-ground novel-view synthesis} comparing ZeroNVS model finetuned on MegaScenes (MS) and our data (Ours), with finetuning improves all metrics.}
\label{tab:zeronvs_results}
\vspace{-0.2in}
\end{table}

\noindent \textbf{Results and Discussions.} Table~\ref{tab:zeronvs_results} presents significant quantitative improvements for single image aerial-to-ground novel-view synthesis. 
From the qualitative results in Figure~\ref{fig:nvs_results}, we see that ZeroNVS (Ours), produces realistic and accurate images that follow the desired poses. In contrast, ZeroNVS (MS), which was finetuned solely on MegaScenes, struggles with such views, highlighting once again the effectiveness of incorporating ground-aerial data into the training process.
We emphasize that this is still a very challenging task, as the viewpoint difference between the reference and target pose is large. The network must learn to retain the underlying scene structure while generating plausible images for unseen parts of the scene and/or demonstrate correct occlusions. 
Our results show that the model somewhat successfully addresses this challenge, but much research remains to be done in this task.

\section{Conclusion}
\label{sec:conclusion}

Despite notable advances in learning-based 3D reconstruction, large-area reconstruction from a sparse mix of drone and ground imagery remains a challenge. As shown over the last decade, adding significant data where little existed before improves the performance of supervised-learning-based networks. The key innovation in our work comes from understanding how geospatial platforms and crowd-sourced imagery can be combined to provide a potentially unlimited amount of data for training large aerial-ground 3D models. Carefully finetuning existing 3D models with our data showed nearly 15$\times$ improvement in camera estimation and registration, which is at the heart of the large-scale reconstruction problem. We hope our hybrid data framework will help spur further research in the area.

In the future, aerial drone views could serve as a bridge between ground and satellite views, where abundant data is widely available. Combining them all could bring us closer to the ambitious goal of planet-scale 3D reconstruction.

\small \noindent \textbf{Acknowledgements:} This work was supported by Intelligence Advanced Research Projects Activity (IARPA) via Department of Interior/Interior Business Center (DOI/IBC) contract number 140D0423C0074. The U.S. Government is authorized to reproduce and distribute reprints for Governmental purposes notwithstanding any copyright annotation thereon. Disclaimer: The views and conclusions contained herein are those of the authors and should not be interpreted as necessarily representing the official policies or endorsements, either expressed or implied, of IARPA, DOI/IBC, or the U.S. Government.

{
    \small
    \bibliographystyle{ieeenat_fullname}
    \bibliography{main}
}

\clearpage
\appendix

\renewcommand{\thesection}{\Alph{section}}
\section*{Appendix}

% \noindent \textcolor{red}{For more 3D interactive visualizations and results, please visit our website! }

\noindent \textbf{Data details:} We use Google Earth Studio (GES, \url{https://earth.google.com/studio}) to render pseudo-synthetic images. Following the method outlined in the main paper, we automatically generate viewpoints for each scene and render the images using GES. Google Earth images are rendered at $1920 \times 1080$ resolution with a randomly varied hFOV between 45\degree and 90\degree to introduce intrinsics diversity. Additionally, images are rendered at different times of day to capture appearance variations. While Google Earth doesn't provide ground-truth depth maps, we have access to ground-truth camera poses. Since we can render the scene from any viewpoint, with dense coverage, depth maps generated using classical MVS are sufficiently accurate to be used as supervision data. 

\noindent \textbf{Training details.} We fine-tune the released DUSt3R/MASt3R checkpoints for 20 epochs using 8 RTX A6000 GPUs with an effective batch size of 32. Following the original setup, for each epoch we sample 100K random image pairs from our AerialMegaDepth dataset, and images are resized so that the largest dimension is 512 pixels. We use the AdamW~\cite{loshchilov2017decoupled} optimizer with a learning rate of $1\text{e-5}$.

\begin{figure}[h!] \centering
    \includegraphics[width=\columnwidth]{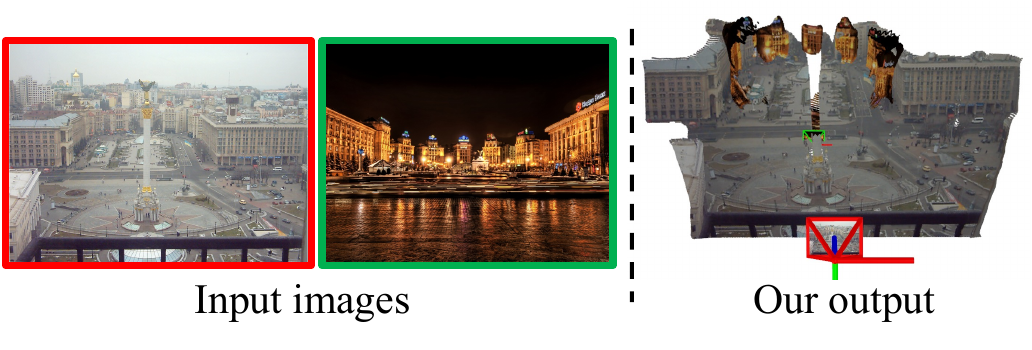}
    \vspace{-0.35in}
    \caption{Our model generalizes well to lighting variations.} \label{fig:lighting_variation}
    \vspace{-0.15in}
\end{figure}

\noindent \textbf{Generalization under varying lighting conditions:} As both Google Earth and MegaDepth images contain lighting variation, our model generalizes reasonably well under these conditions as shown in Figure~\ref{fig:lighting_variation} (image pair from WxBS~\cite{wxbs}).

\vspace{-0.10in}

\noindent \textbf{Expanded evaluation:} In addition to the ground-aerial setting evaluated in the main paper, we also include results for \textit{ground-ground} and \textit{aerial-aerial} configurations in Table~\ref{tab:pairwise_results_supp}. This table shows that DUSt3R and MASt3R maintain strong performance on ground-ground and aerial-aerial pairs, which typically have higher visual overlap. This demonstrates that fine-tuning with our varying-altitude data does not degrade performance on these easier, similar-viewpoint settings.

\definecolor{first}{rgb}{1.0, .83, 0.3}
\definecolor{second}{rgb}{1.0, 0.93, 0.7}
\def \first {\cellcolor{first}}
\def \second {\cellcolor{second}}

\begin{table*}[h]
\centering
\resizebox{0.95\textwidth}{!}{%
\begin{tabular}{c|l|ccc|ccc|ccc}
\toprule
\multicolumn{1}{c|}{\multirow{2}{*}{Eval. Split}} & \multicolumn{1}{c|}{\multirow{2}{*}{Method}} & \multicolumn{3}{c|}{\emph{Camera Rotation Accuracy}} & \multicolumn{3}{c|}{\emph{Camera Translation Accuracy}} & \multicolumn{3}{c}{\emph{3D Pointmap Accuracy}} \\ \cline{3-11} 
\multicolumn{1}{c|}{}                            & \multicolumn{1}{c|}{}                        & $\RRA$@5\degree              & $\RRA$@10\degree             & $\RRA$@15\degree            & $\RTA$@5\degree             & $\RTA$@10\degree            & $\RTA$@15\degree           & $\delta$@0.5m           & $\delta$@1m          & $\delta$@2m          \\ \midrule

\multirow{7}{*}{\rotatebox[origin=c]{90}{ground-ground}}         
& LoFTR$^{*}$~\cite{sun2021loftr}                                & 93.60 & 95.97 & 96.92 & 78.20 & 88.63 & 92.65 & - & - & - \\ 
& SP+SG$^{*}$~\cite{superglue}                                & 92.89 & 95.50 & 95.73 & 72.75 & 88.39 & 93.60 & - & - & - \\ \cline{2-11}
& MASt3R~\cite{mast3r_arxiv24} (released)                                       & 86.49 & 92.89 & 95.02 & 53.32 & 76.07 & 81.99 & 43.42 & 64.23 & 74.82 \\  
& DUSt3R~\cite{dust3r_cvpr24} (released)                                   & 90.52 & 95.26 & 96.45 & \first \textbf{61.61} & 77.96 & 83.18 & 49.11 & 66.70 & 73.93 \\
& \textbf{MASt3R + PSynth (Ours)}                                & 91.94 & 95.02 & 96.92 & 45.50 & 69.91 & 82.23 & \second 51.58 & \second 68.43 & \second 75.74 \\
& \textbf{DUSt3R + PSynth (Ours)}                                & \second 92.42 & 95.73 & 96.92 & \second 57.82 & \first \textbf{78.20} & 83.41 & 50.99 & 67.63 & 74.74 \\
& \textbf{MASt3R + Hybrid (Ours)}                                & 91.71 & \second 96.21 & \second 97.39 & 48.34 & 69.67 & 77.25 & \first \textbf{52.32} & \first \textbf{69.23} & \first \textbf{76.31} \\           
 & \textbf{DUSt3R + Hybrid (Ours)}                                & \first \textbf{94.55} & \first \textbf{97.63} & \first \textbf{98.10} & 55.69 & \second 77.96 & \first \textbf{85.31} & 50.18 & 67.98 & 75.63 \\
                                                                           
                                                 \midrule

\multirow{7}{*}{\rotatebox[origin=c]{90}{aerial-aerial}}     
& LoFTR$^{*}$~\cite{sun2021loftr}                                & 96.00 & 96.62 & 96.62 & 92.92 & 95.38 & 96.31 & - & - & - \\ 
& SP+SG$^{*}$~\cite{superglue}                                & 95.08 & 95.69 & 95.69 & 92.00 & 95.69 & 95.69 & - & - & - \\ \cline{2-11}
& MASt3R~\cite{mast3r_arxiv24} (released)                                       & \first \textbf{99.69} & 100.00 & 100.00 & 77.23 & 91.69 & 96.31 & 19.10 & 45.37 & 68.09 \\
& DUSt3R~\cite{dust3r_cvpr24} (released)                                       & \second 98.77 & 100.00 & 100.00 & 55.08 & 90.15 & 95.38 & 11.24 & 37.68 & 64.78 \\
& \textbf{MASt3R + PSynth (Ours)}                                & 98.46 & 100.00 & 100.00 & \first \textbf{84.92} & \first \textbf{96.31} & \second 96.31 & \second 26.65 & \second 54.99 & \second 74.33 \\  
& \textbf{DUSt3R + PSynth (Ours)}                                & 98.46 & 100.00 & 100.00 & \second 84.92 & 94.46 & 96.31 & 16.07 & 44.40 & 70.37 \\
& \textbf{MASt3R + Hybrid (Ours)}                                & 98.46 & \first \textbf{100.00} & \first \textbf{100.00} & 80.62 & \second 95.08 & \first \textbf{96.31} & \first \textbf{27.93} & \first \textbf{57.76} & \first \textbf{75.55} \\
 & \textbf{DUSt3R + Hybrid (Ours)}                                & 98.46 & \second 100.00 & \second 100.00 & 80.62 & 92.31 & 95.08 & 14.14 & 41.05 & 66.25 \\
                                                  
 \bottomrule 
\end{tabular}
}

\caption{\textbf{Expanding on the results from Table 1 of the main paper, we include evaluations for \textit{ground-ground} and \textit{aerial-aerial} settings.} While Table 1 emphasizes the significant improvements achieved in the challenging ground-aerial setting through fine-tuning with our data, this table shows that DUSt3R and MASt3R also maintain strong performance on ground-ground and aerial-aerial pairs, which typically have higher visual overlap. This demonstrates that fine-tuning with our varying-altitude data does not degrade performance on these easier, similar-viewpoint cases.}

\label{tab:pairwise_results_supp}
\end{table*}

\end{document}